# Reinforcement learning reward function in unmanned aerial vehicle control tasks

**M S Tovarnov[1], N V Bykov[1,2], ***

[1] Bauman Moscow State Technical University (BMSTU), 5/1, 2-nd Baumanskaya Str, Moscow 105005, Russia

[2] Semenov Federal Research Center for Chemical Physics of the Russian Academy of Sciences, 4, Kosigin Str., Moscow 119991, Russia

* Corresponding author: bykov@bmstu.ru

**Abstract.** This paper presents a new reward function that can be used for deep reinforcement learning in unmanned aerial vehicle (UAV) control and navigation problems. The reward function is based on the construction and estimation of the time of simplified trajectories to the target, which are third-order Bezier curves. This reward function can be applied unchanged to solve problems in both two-dimensional and three-dimensional virtual environments. The effectiveness of the reward function was tested in a newly developed virtual environment, namely, a simplified two-dimensional environment describing the dynamics of UAV control and flight, taking into account the forces of thrust, inertia, gravity, and aerodynamic drag. In this formulation, three tasks of UAV control and navigation were successfully solved: UAV flight to a given point in space, avoidance of interception by another UAV, and organization of interception of one UAV by another. The three most relevant modern deep reinforcement learning algorithms, Soft actor-critic, Deep Deterministic Policy Gradient, and Twin Delayed Deep Deterministic Policy Gradient were used. All three algorithms performed well, indicating the effectiveness of the selected reward function.

## 1. Introduction

Recent advances in information technology and artificial intelligence have had a significant impact on intelligent autonomous systems. These advances also affected unmanned aerial vehicles (UAVs). The initial proliferation of UAVs began with their use in filmmaking, offering a new level of creativity in photography and video production. UAVs have also been employed in some commercial and security fields [1]. UAVs also play significant roles in promising concepts of "smart cities": distributing the internet, delivering parcels, monitoring and controlling the flow of urban traffic, identifying threats and ensuring the security of a "smart city" [2].

Currently, there are tasks where serious automation of vehicle control is required in all areas of UAV application: route planning, navigation and direct control. In other words, UAVs must be able to perform planned missions in difficult and unexpected situations without human control. Despite the existing variety of well-established methods for solving such problems, some approaches are being actively developed thanks to the tremendous progress in the fields of machine learning and artificial intelligence. One such approach is reinforcement learning (RL) [3], which aims to find the optimal control function. In areas related to UAVs, RL algorithms are also widely used [3–12].

RL has an interactive intelligent agent with a clearly defined goal. An agent in the context of RL is an entity that performs any action in the environment, taking into account its current state. The main goal of any RL algorithm is to allow the agent to quickly learn optimal behavior that accurately performs the assigned task and leads to the highest value of the reward [13].

Rewards are numerical values that the environment gives to the agent in response to a pair of (state, action). Rewards quantify the actions of the agent. The reward function is the cornerstone of RL-related challenges. The speed of learning and the fundamental possibility of solving the problem depend on its choice [14]. Despite the existing successes of this approach, RL has a number of features that greatly complicate its application in real problems: high computational complexity as well as the need for a careful selection of the reward function. Reinforcement learning is a type of supervised machine learning. However, training data are obtained as a result of the agent's interaction with the environment. Given the high computational complexity of RL algorithms, it is almost impossible to perform training on a real object (UAV). Therefore, training is performed in a virtual environment.

Many tasks of automation of control and navigation of UAVs have common formulations and subtasks, among which three main tasks can be distinguished:

1) The UAV must fly through certain points in space [4,7,8,10–12].

2) The UAV must fly to a certain point in space, avoid obstacles, or avoid interception by another UAV [15–18].

3) The UAV should intercept another UAV [6,9,19].

All of these tasks can be reduced to the following: the UAV must reach a certain point in space while having a certain velocity vector there. The destination point position and velocity vector can be time dependent. For example, in the task of intercepting a UAV, the destination point coincides with the current position of the UAV that needs to be intercepted.

The construction of the reward function proposed in this paper is based on the provision according to which the proximity of an agent to its target is estimated by the flight time to it. This reward function rewards the agent's actions that shorten this time and penalizes the agent for actions that increase it. This approach is applicable in both three-dimensional and two-dimensional space. An estimate of the flight time to the target can be obtained by constructing a simplified trajectory, which should simultaneously take into account the velocity and maneuverability of the UAV. This can be achieved using Bezier curves [20].

Despite the fact that the studies in this paper are performed on a well-known mathematical model of the environment, as an approach to training the agent, algorithms were used that do not use a priori knowledge about the environment and perceive it as a "black box". Such algorithms are generally called model free [21]. This approach to learning makes it possible to solve increasingly complex problems using more complex environmental models.

On the one hand, the environmental model itself is simple to calculate; on the other hand, it is extremely dynamic and nonlinear from the point of view of UAV control. Therefore, the task of training an agent was solved using the most universal approach, namely, deep reinforcement learning [13], where a multilayer neural network acts as a function for choosing an agent's action. A wide variety of approaches are available for the formation and use of learning algorithms for such neural networks [3,13,21].

In this paper, agents are trained in a two-dimensional environment because there are no fundamental differences with respect to the three-dimensional formulation. However, the two-dimensional formulation allows one to reduce the dimension of the space of states and actions of the agent, which reduces the training time.

## 2. Reward function

Assume that at time $t = t_0$, the UAV is at point **A** and has a velocity $\mathbf{v}_A$. In addition, at time $t = t_1$, it should be at point **D** and have a velocity $\mathbf{v}_D$. The trajectory of such a UAV is described by a third-order Bezier curve, which is constructed from four points in space: **A**, **B**, **C** and **D** (figure 1). In this case, points **B** and **C** are determined based on the boundary conditions for velocities.

The result is a parametric curve:

$$\mathbf{r}(\tau) = (1-\tau)^3 \mathbf{A} + \tau(1-\tau)^2 \mathbf{B} + \tau^2(1-\tau)\mathbf{C} + \tau^3 \mathbf{D},$$

$$\tau(t) = \frac{t - t_0}{t_1 - t_0} \in [0,1], \quad \mathbf{B} = \mathbf{A} + \mathbf{v}_A \frac{t_1 - t_0}{3}, \quad \mathbf{C} = \mathbf{D} - \mathbf{v}_D \frac{t_1 - t_0}{3}.$$

Flight along this trajectory assumes that the vectors of velocity **v** and acceleration **a** at each moment of time are determined as follows:

$$\mathbf{v}(\tau) = \frac{3}{t_1 - t_0}\left[(1-\tau)^2(\mathbf{B} - \mathbf{A}) + 2\tau(1-\tau)(\mathbf{C} - \mathbf{B}) + \tau^2(\mathbf{D} - \mathbf{C})\right],$$

$$\mathbf{a}(\tau) = \frac{6}{(t_1 - t_0)^2}\left[(\tau - 1)(\mathbf{B} - \mathbf{A}) + (1 - 2\tau)(\mathbf{C} - \mathbf{B}) + \tau(\mathbf{D} - \mathbf{C})\right].$$

The above relations define a one-parameter family of curves, each of which is determined by a single parameter - the time of flight between points **A** and **D**: $\Delta t = t_1 - t_0$. By specifying $\Delta t$, different trajectories can be obtained. The hodographs of the UAV velocity vectors on these trajectories are second-order Bezier curves. The UAV acceleration hodographs on these trajectories are first-order Bezier curves.

The shorter the time $\Delta t$, the greater the maximum velocities and accelerations on the trajectory. The choice of a particular trajectory is determined by constraints on the values of the maximum velocities and accelerations. Specifically, one can find such a minimum time $\Delta t^{\min}$ at which the velocities and accelerations on the trajectory will fully satisfy their constraints.

Thus, the reward function can be written as follows:

$$r_i = \Delta t_{i-1}^{\min} - \Delta t_i^{\min},$$

where $r_i$ is the reward to the agent at the $i$-th step during the transition from state $s_{i-1}$ to state $s_i$; $\Delta t_{i-1}^{\min}$ and $\Delta t_i^{\min}$ are estimates of the time of the fastest satisfactory trajectories in states $s_{i-1}$ and $s_i$, respectively. Accordingly, the agent is positively rewarded if it was able to reduce the time to achieve the goal and is fined otherwise.

## 3. Mathematical model of the environment

The agent controls the UAV of the quadrocopter type by sending signals to the engines. At equal time intervals $t_s$, the agent issues a new value of the control vector, which determines the output power of the engines on each propeller. Time $t_s$ is determined by the hardware capabilities of the UAV control systems. The values of the components of the steering vector are continuous and vary in the range [0, 1]. If the component is zero, then the corresponding engine does not produce thrust. If the component is equal to 1, thrust is maximum. This formulation of control is a complicated control model for the agent given that, in addition to determining the optimal path, it is also responsible for stabilizing the position, velocity, orientation in space and angular velocities of the UAV.

The UAV is assumed to be a rigid body moving in a two-dimensional coordinate system under the action of the traction forces of propellers $\mathbf{F}_1$, $\mathbf{F}_2$, gravity $m\mathbf{g}$, and air drag force $\mathbf{X}$, which is directed opposite to the UAV velocity vector **v**. Forces $\mathbf{F}_1$ and $\mathbf{F}_2$ are always directed perpendicular to the UAV axis. At point **A** (figure 2) is the center of mass of the UAV, which seeks to fly to point **D**. Additionally, in some tasks (see below) at point **M,** there may be an aircraft that seeks to intercept the UAV.

The mathematical model of UAV dynamics is a system of differential equations describing the plane motion of a rigid body with constant mass $m$ as follows:

$$\frac{dv_x}{dt} = \frac{F_{\max}(H)(a_1 + a_2)}{m}\sin(-\beta) - \frac{C_x S}{m}\frac{\rho(H)|\mathbf{v}|^2}{2}\cos(\alpha + \beta),$$

$$\frac{dv_y}{dt} = \frac{F_{\max}(H)(a_1 + a_2)}{m}\cos(\beta) - g - \frac{C_x S}{m}\frac{\rho(H)|\mathbf{v}|^2}{2}\sin(\alpha + \beta),$$

$$\frac{d\omega}{dt} = \frac{hF_{\max}(H)(a_2 - a_1)}{I_z}, \quad \frac{dx}{dt} = v_x, \quad \frac{dH}{dt} = v_y, \quad \frac{d\beta}{dt} = \omega,$$

$$|\mathbf{F}_1| = a_1 F_{\max}(y), \quad |\mathbf{F}_2| = a_2 F_{\max}(y), \quad a_i \in [0,1],$$

where $x$ and $H$ are coordinates of the center of mass of the UAV; $v_x$ and $v_y$ are components of the velocity vector $\mathbf{v}$; $C_x$ is the drag coefficient; $S$ is the characteristic area of the UAV; $\omega$ is the angular velocity of rotation of the UAV; $\alpha$ is the angle of attack of the UAV; $\beta$ is the angle of inclination of the UAV axis; $|\mathbf{v}|$ is the modulus of the UAV velocity vector; $\rho(y)$ is a function of the atmospheric density on the height, which can be found in [22]; $\mathbf{g}$ is the acceleration of gravity; $h$ is the arm of forces $\mathbf{F}_1$ and $\mathbf{F}_2$ relative to the center of mass; $I_z$ is the moment of inertia of the UAV; $F_{\max}(y)$ is the maximum rotor thrust at height $y$; and $a_i$ is the relative thrust on the $i$-th propeller.

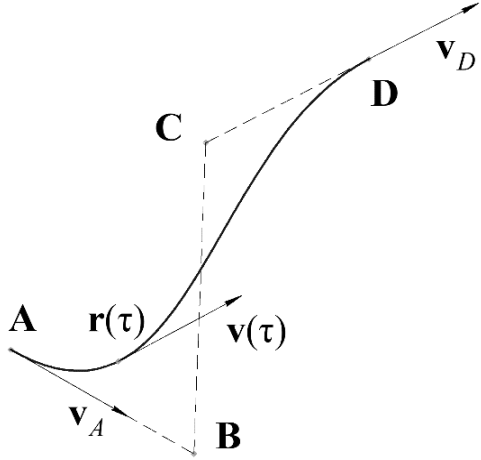
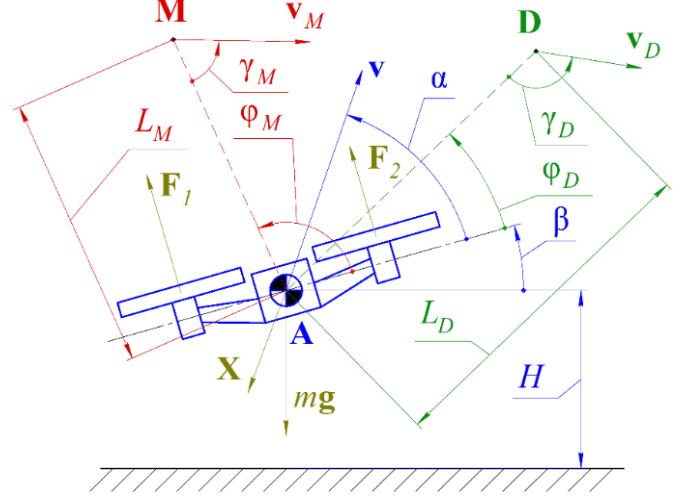

**Figure 1.** UAV trajectory construction.      **Figure 2.** Diagram of the forces acting on the UAV.

The maximum thrust of the UAV propeller is related to the density of the ambient air as follows [23]:

$$\frac{F_{\max}(H)}{F_{\max}(0)} = \sqrt[3]{\frac{\rho(H)}{\rho(0)}}.$$

At each step, the agent receives a vector of observations of the current environment and the state of the UAV as input data:

$$\mathbf{s} = \left(H, \omega, \alpha, \beta, |\mathbf{v}|, L_D, |\mathbf{v}_D|, \varphi_D, \gamma_D, L_M, |\mathbf{v}_M|, \varphi_M, \gamma_M\right)^T,$$

where $H$ is the height above the surface; $L_D$ is the distance to the target; $|\mathbf{v}_D|$ is the modulus of the vector of the required UAV velocity; $\varphi_D$ is the angle between the line of sight **AD** and the UAV axis; $\gamma_D$ is the angle between the line of sight **AD** and the vector $\mathbf{v}_D$; $L_M$ is the distance to the opposing aircraft; $|\mathbf{v}_M|$ is the modulus of the velocity vector of the opposing aircraft; $\varphi_M$ is the angle between the line of sight **AM** and the UAV axis; and $\gamma_M$ is the angle between the line of sight **AM** and the vector $\mathbf{v}_M$. In tasks where there is no opposing aircraft, the last four components of the vector $\mathbf{s}$ are equal to zero. As an output at each step, the agent determines the vector of relative thrust $\mathbf{a} = (a_1, a_2)$.

## 4. Scenarios and algorithms for training agents

The agents were trained in the following scenarios: (1) UAV flight to a random fixed point in space; (2) UAV flight to a random fixed point in space, whereas the UAV must avoid interception by more maneuverable aircraft, which has an ideal control system; (3) interception of a UAV by another UAV (both agents are trained).

All tasks are modeled in a rectangular area $x \subset [-5000, 5000]$, $H \subset [0, 3000]$. If the UAV flies out of its borders, then the current episode ends, and the agent is penalized by 100 points. If the module of the UAV's angular velocity exceeds 20 rad/s, then the current episode ends, and the agent is penalized by 100 points.

In the calculations, the following model values of the UAV parameters were used: $m = 5$ kg; $F_{\max}(0) = 42$ N; $I_z = 1.25$ kg·m²; $C_xS = 0.2$ m²; $h = 0.5$ m; and $t_s = 0.05$ s. Based on these parameters, the maximum velocity that the UAV is able to develop is 35 m/s, and the maximum available overload is 16.8 m/s². These values roughly correspond to the real characteristics of commercial UAVs.

*4.1. UAV flight to a random fixed point in space*

Each new episode starts with two random points **A** and **D** placed in a rectangular area $x \subset [-4000, 4000]$, $H \subset [0, 2500]$. A UAV is initially placed at point **A** and has a random state: a velocity vector (angle from $-\pi$ to $\pi$, modulus from 0 to 2 m/s), angle of inclination of the UAV axis (from $-10°$ to $+10°$) and angular velocity (from $-0.01$ to $+0.01$ rad/s). The UAV needs to fly to point **D**. The velocity vector $\mathbf{v}_D$, which the UAV should have at point **D**, is randomly generated (angle from $-\pi$ to $\pi$, modulus from 1 to 13 m/s). Furthermore, at each step, the agent generates control signals based on current observations until the end of the episode. The condition for the successful end of the episode is considered if the UAV is in the vicinity of the required point (the distance to the point is less than $L_{stop} = 10$ m). The final reward in this case is calculated as follows:

$$r_{final} = r_n + 100 \frac{2 L_{stop}}{|\mathbf{v}_D + \mathbf{v}_n|} \frac{1}{\Delta t_n^{min}},$$

where $r_n$ is the reward at the last step, excluding the end of the episode, and $\mathbf{v}_n$ is the UAV velocity at the last step. The closer the UAV velocity vector is to the final required velocity vector $\mathbf{v}_D$, the greater the reward.

*4.2. UAV flight to a random fixed point in space, while the UAV must avoid interception by more maneuverable aircraft, which has an ideal control system*

In each new episode, the UAV and its target (destination point **D**) are generated in the same manner as noted in the case of a previous scenario of a UAV flight to a random fixed point in space. In this scenario, an aircraft is additionally present, and its task is to intercept the UAV. Such an interceptor is a point in space that moves with a constant modulus velocity $\mathbf{v}_M$ and can perform maneuvers with lateral overload $\mathbf{a}_M$.

The interceptor flies to the anticipatory point $\mathbf{A}_t$, the position of which is calculated based on the current position and velocity of the UAV:

$$\mathbf{A}_t = \mathbf{A} + \mu \mathbf{v}, \quad \mu = \lambda \frac{|\mathbf{A} - \mathbf{M}|}{|\mathbf{v}_M|}, \lambda \in [0,1],$$

where $\lambda$ is the coefficient of proportionality. When $\lambda = 0$, the interceptor pursues the UAV using the "pure pursuit" method. When $\lambda = 1$, it pursues the proportional approach method. The interceptor maneuvers toward the target only if the point $\mathbf{A}_t$ is outside the line of sight, which is limited by the angle $\varepsilon_M$. The mathematical model of the movement of such an interceptor is reported in the following equations:

$$\frac{dx}{dt} = |\mathbf{v}_M| \cos\theta_M, \quad \frac{dH}{dt} = |\mathbf{v}_M| \sin\theta_M,$$

$$\frac{d\theta_M}{dt} = \begin{cases} 0, \text{ if } \angle(\mathbf{v}_M, \mathbf{A}_t - \mathbf{M}) \leq \varepsilon_M/2, \\ a_M/|\mathbf{v}_M|, \text{ if } \angle(\mathbf{v}_M, \mathbf{A}_t - \mathbf{M}) > \varepsilon_M/2 \text{ sign}(\mathbf{v}_M \times (\mathbf{A}_t - \mathbf{M})), \\ -a_M/|\mathbf{v}_M|, \text{ otherwise,} \end{cases}$$

where $\theta_M$ is the pitch angle of the interceptor, and sign is a signum function.

At the beginning of each new episode, an interceptor with random characteristics is placed at a random point in some neighborhood around the UAV. The neighborhood is determined by an estimate of the intercept time so that the interceptor has the theoretical ability to intercept the UAV before it reaches its destination point. The radius of the neighborhood is determined by the formula: $R = 0.9 \Delta t_0^{min} |\mathbf{v}_M|$.

The conditions for the end of the episode are similar to those in scenario 1. In addition, one more condition is added. Specifically, if the interceptor approaches the UAV closer than $L_{stop} = 10$ m, then it is considered that it has successfully intercepted the UAV, and the current episode ends. In this case, the agent is penalized 100 points.

*4.3. Interception of a UAV by another UAV*

In each new episode, the UAV and its target (destination point **D**) are generated in the same manner as noted in the case of a first scenario of a UAV flight to a random fixed point in space. In this scenario, there is another similar UAV, and its task is to intercept the first UAV. At the initial moment of time, this interceptor is located in a certain vicinity of destination point **D** of the first UAV. Each UAV is controlled by its own agent. To calculate the reward function for an interceptor agent, the same approach is used as for a stationary target. However, in this case, the final destination of the interceptor moves. Specifically, the position of the destination point at each step of the episode coincides with the current position of the first UAV, and the vector of the required velocity in it is related in magnitude to the current velocity of the first UAV based on the following relationship: $|\mathbf{v}'_A| = 1.2|\mathbf{v}|$. The vector $\mathbf{v}'_A$ is directed along the line of sight **AM**.

In this scenario, two agents are trained simultaneously. The first agent is trained according to the rules of scenario 2. The interceptor agent is trained according to the rules of scenario 1 with the exception that the reward function is simplified. Here, in case of successful interception, the agent receives 100 points.

*4.4. Agent training algorithms*

The reward function was tested on modern RL algorithms that work with a continuous space of actions and have shown their effectiveness in UAV control and navigation tasks:

1) Soft actor-critic (SAC) [24]. This modern method shows good results in problems with continuous action spaces, whereas its requirements for setting hyperparameters is not very stringent. The method combines the advantages of good convergence of on-policy methods and the ability to reuse data for training as in off-policy methods.

2) Deep Deterministic Policy Gradient (DDPG) [25]. This method was one of the first to be able to reuse data for training in problems with a continuous space of actions.

3) Twin Delayed Deep Deterministic Policy Gradient (TD3) [26]. This method is an improved version of DDPG. The main changes were aimed at improving convergence and reducing the requirements for tuning hyperparameters.

## 5. Results and discussion

The algorithms were used independently in each scenario. The training lasted the same amount of time for each scenario and each algorithm. The neural network architectures responsible for the choice of agent actions used for all algorithms were taken to be the same. The network is two-layer and consists of fully connected layers with rectified linear unit (ReLU) activation functions with 400 neurons in the first layer and 300 neurons in the second layer. Learning progress was assessed by final agent rewards in each episode. The results (figures 3–5) show the curves of the mathematical expectation of the reward at the end of the episode and the boundaries of the 95% percentile for a sample of the last 200 episodes of training.

Figure 3 shows the progress of the training models in the first scenario. All three algorithms, after 2000 episodes of the script, learned how to perform the task. The average performance of subsequent episodes improved due to an increase in the proportion of favorable episodes and an increase in the final rewards in them.

The results of agent training in the second scenario (figure 4) reveal the following: UAVs controlled by the agent have not learned with 100% probability to avoid the ideal interceptor. However, the share of successful executions of tasks exceeds the share of successful interceptions. For the flight to the destination point at a certain angle with a certain velocity, as the results showed, the primary task for agents is to avoid interception and not to match the velocity vector at the end point to the required value.

The third scenario, in which one UAV of the first agent must reach a fixed point in space and the second agent controls an interceptor UAV and must intercept the first UAV, showed the following results (figure 5). The interceptor performs its role better than the interceptor from the second scenario; however, it fails to achieve 100% interception.

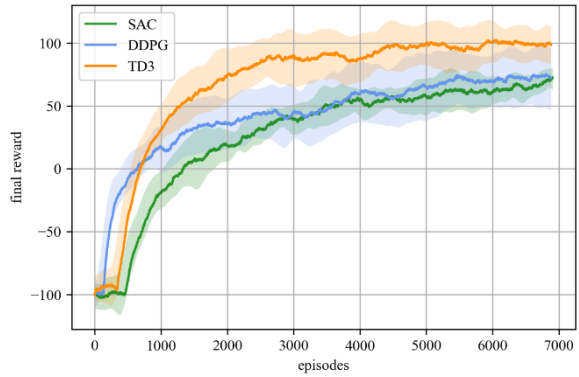

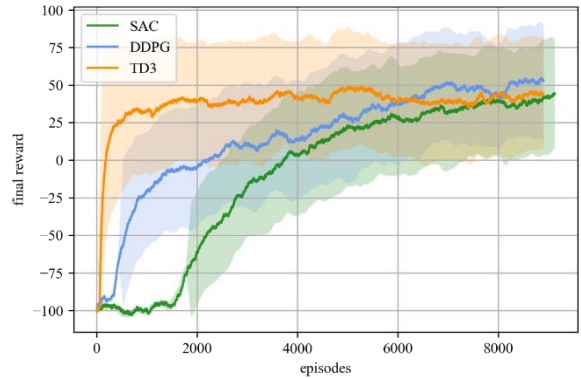

**Figure 3**. Progress of agent training in the first scenario.

**Figure 4.** Progress of agents training in the second scenario.

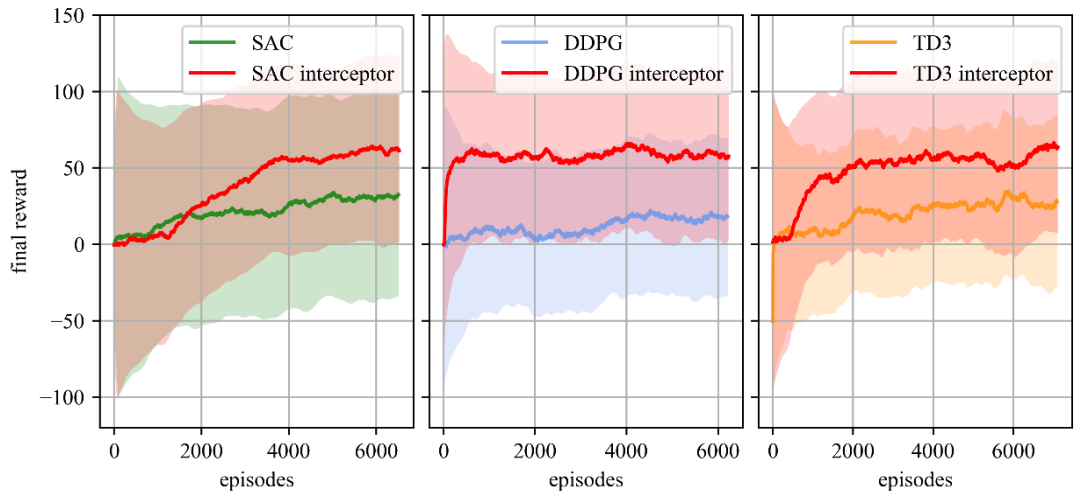

**Figure 5.** Progress of agents training in the third scenario.

The percentage of successful interceptions exceeds the percentage of successful missions with the first UAV.

## 6. Conclusions

We have shown that the reward function, which is based on a simplified estimate of the flight time to the target using third-order Bezier curves, can be used in modern deep reinforcement learning algorithms to solve the following problems: UAV flight to a random fixed point in space; UAV flight to a random fixed point in space in the presence of an ideal interceptor; and interception of one UAV by another UAV. Such a reward function can be applied without changes to solve problems of control and navigation of UAVs in both two-dimensional and three-dimensional virtual environments. The effectiveness of the proposed reward function was tested using three modern reinforcement learning algorithms: Soft actor-critic, Deep Deterministic Policy Gradient, and Twin Delayed Deep Deterministic Policy Gradient. All three algorithms showed good results, which indicates the effectiveness of such a reward function for solving the problems considered.

This approach is planned to be further developed in more complex and stochastic virtual environments, which take into account more of the features inherent in a real UAV flight. However, this process will require incomparably more computing resources.

## Acknowledgments

The reported study was funded by the Russian Foundation for Basic Research (RFBR, project number 19–29–06090 mk).